\pdfoutput=1

\documentclass[11pt]{article}

\usepackage{ACL2023}

\usepackage{times}
\usepackage{latexsym}

\usepackage[T1]{fontenc}

\usepackage[utf8]{inputenc}

\usepackage{microtype}

\usepackage{inconsolata}

\title{Symbolic Prompt Program Search: A Structure-Aware Approach \texorpdfstring{\\}{} to Efficient  Compile-Time Prompt Optimization}
\usepackage{graphicx}
\usepackage{subcaption}
\usepackage{booktabs}
\usepackage{hyperref}

\usepackage{svg}
\usepackage{setspace}
\usepackage{amsmath}
\usepackage{fvextra}
\fvset{breaklines=true}
\usepackage{amssymb}
\usepackage{mathtools}
\usepackage{amsthm}
\usepackage{tikz}
\usepackage{url}
\usepackage{times}
\usepackage{latexsym}
\usepackage{algorithm}
\usepackage{fancyvrb}
\usepackage{tabularx}
\usepackage{algpseudocode}
\usepackage{amsmath}
\usepackage{textcomp}
\usepackage{bm}
\usepackage[linguistics]{forest}
\usepackage{xspace}
\usepackage[T1]{fontenc}
\usepackage{soul}
\usepackage{comment}
\usepackage{multicol}
\usepackage{multirow}
\usepackage{tabularx}
\usepackage[capitalize,noabbrev]{cleveref}

\algrenewcommand\alglinenumber[1]{\tiny #1:}
\usepackage[textsize=tiny]{todonotes}

\newcommand{\ours}{\textsc{Sammo}\xspace}
\newcommand{\repo}{\url{https://anonymous.4open.science/r/sammo-4003}\xspace}
\algnewcommand\algorithmicswitch{\textbf{switch}}
\algnewcommand\algorithmiccase{\textbf{case}}
\algnewcommand\algorithmicassert{\texttt{assert}}
\algnewcommand\Assert[1]{\State \algorithmicassert(#1)}%
\definecolor{lightergray}{gray}{0.95}
\algdef{SE}[SWITCH]{Switch}{EndSwitch}[1]{\algorithmicswitch\ #1\ \algorithmicdo}{\algorithmicend\ \algorithmicswitch}%
\algdef{SE}[CASE]{Case}{EndCase}[1]{\algorithmiccase\ #1}{\algorithmicend\ \algorithmiccase}%
\algtext*{EndSwitch}%
\algtext*{EndCase}%
\algnewcommand\algorithmicforeach{\textbf{for each}}
\algdef{S}[FOR]{ForEach}[1]{\algorithmicforeach\ #1\ \algorithmicdo}

\theoremstyle{plain}

\theoremstyle{definition}

\theoremstyle{remark}

\ifdefined \argmax
\else \DeclareMathOperator*{\argmax}{arg\,max}
\fi

\begin{document}

\author{
Tobias Schnabel \and Jennifer Neville\\ 
Microsoft Research, Redmond, USA
\\
\texttt{\{toschnab,jenneville\}@microsoft.com}
}

\maketitle

\begin{abstract}
In many modern LLM applications, such as retrieval augmented generation, prompts have become programs themselves. In these settings, prompt programs  are repeatedly called with different user queries or data instances. A big practical challenge is optimizing such prompt programs. Recent work has mostly focused on either simple prompt programs or assumed that the general structure of a prompt program is fixed. 

We introduce SAMMO, a framework to perform \emph{symbolic} prompt program search for {\em compile-time} optimizations of prompt programs. SAMMO represents prompt programs on a symbolic level which allows for a rich set of transformations that can be searched over during optimization. We show that SAMMO generalizes previous methods and improves the performance of complex prompts on (1) instruction tuning, (2) RAG pipeline tuning, and (3) prompt compression, across several different LLMs.
We make all code available open-source at \repo.
\end{abstract}

\section{Introduction}

\begin{figure*}[!tb]
    \centering
\includegraphics[width=0.99\textwidth]{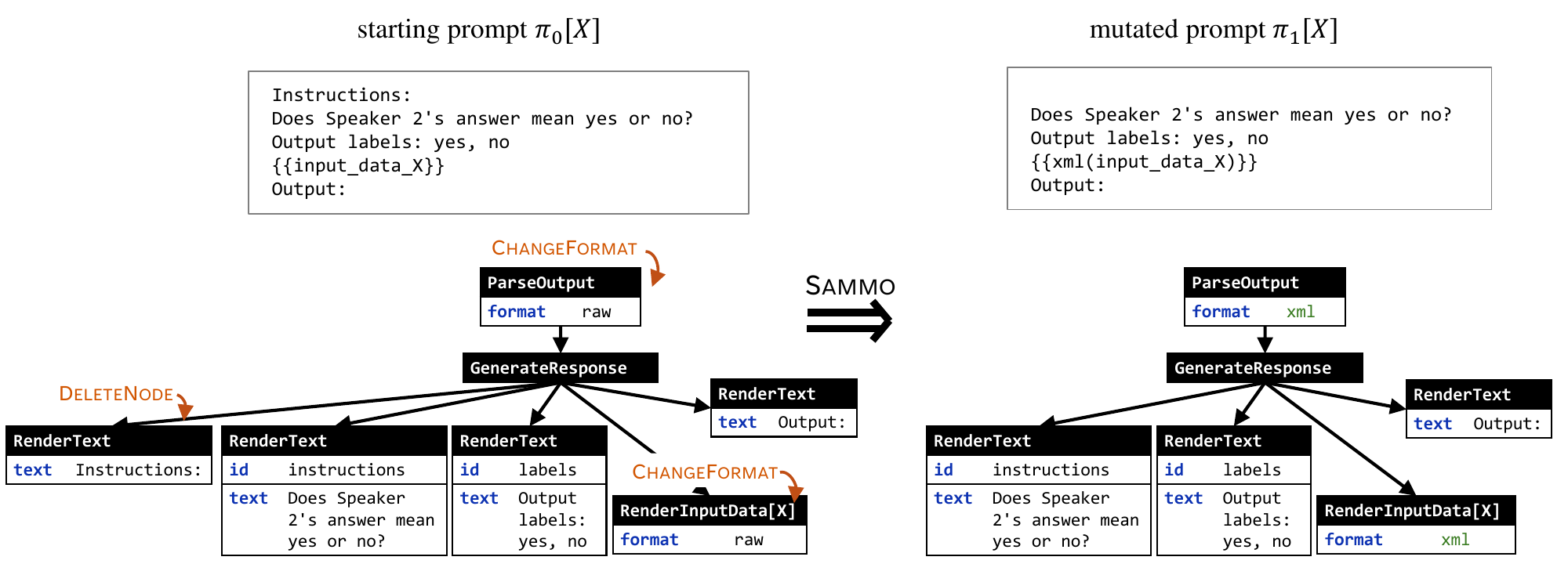}
    \caption{Left: Symbolic prompt program (SPP) for a binary classification task, where each node is a function with attributes and dependencies (children). The example also shows how the SPP allows for structural changes (e.g., \textsc{DeleteNode}) and attribute-based changes (e.g., \textsc{ChangeFormat}) which, after applying, result in the mutated prompt (Right). These enable \ours to 
    explore a large set of possible prompt candidates automatically.}
    \label{fig:metaprompt}
\end{figure*}

With the recent development of large language models (LLMs) such as Mixtral 8x7B~\cite{jiang2024mixtral} or GPT-4, it is now possible to provide LLMs with longer inputs including richer context and more detailed instructions. As a result, the complexity of prompts has increased, including longer strings of instructions, demonstrations or examples, and specification of more structured output. These lengthy instructions are often reused as part of larger prompts, where static information (e.g., instructions) is combined with input-dependent information (e.g., user queries, retrieved documents in RAG systems) at runtime. 
This has led to prompts being regarded as programs themselves, where each part is the result of other subprograms~\cite{khattab2023dspy}. 

As LLM performance depends on a variety of factors such as paraphrasing, formatting and ordering~\cite{sclar2023quantifying,liu2024lost}, a key challenge to solve in practice is automatically finding the best prompt program for a given task and model. In this paper, we focus on \emph{compile-time optimization} of prompt programs under \emph{black-box LLM} access. Compile-time optimization is carried out only once before deployment and has two large advantages: (i) we can amortize the optimization cost over multiple uses of the prompt program and (ii) keep the existing run-time architecture unchanged. This is different from run-time optimization approaches, e.g., token pruning~\cite{jiang2023llmlingua} which are called every time before inference and thus have non-amortizable costs.  

A lot of work on prompt optimization has focused on simple prompt programs where the program consists of a single string (which is subsumed by our representation in Section~\ref{sec:metaprompts}), mostly using structure-unaware operations such as paraphrasing \cite{chen2023instructzero,pryzant2023automatic,zhou2023large}. Another line of work has focused on optimizing prompts with more complex programmatic structure, but still considers only \emph{static} prompt structures, e.g., DSpy~\cite{khattab2023dspy}. Initial work in this direction focused on applying textual mutators to different prompt components \cite{fernando2023promptbreeder, ye2023prompt, khattab2023dspy}. Subsequent work considered both textual mutation and hyperparameter selection \cite{sclar2023quantifying}. 
However, with increasing complexity of prompt structure, many prompt optimization techniques are no longer applicable and a new approach is needed that is able to optimize complex prompt programs with their hyperparameters. 

To address this, we introduce \ours, a general purpose framework for compile-time optimizations of prompt programs. Different from existing approaches \ours represents prompt programs as \emph{symbolic prompt programs} (SPP) which are abstract program graphs that can be changed arbitrarily. This allows \ours to efficiently search through the space of valid prompt programs by automatically generating new promising candidates through mutations. Moreover, \ours allows to both represent the internal structure of prompts (e.g., sections) as well as program structure (e.g., calling a retriever) in a unified fashion. By writing custom mutation operators, practitioners can encode their domain knowledge and guide the search process. \ours naturally extends previous prompt programming approaches such as DSpy~\citep{khattab2023dspy} and encompasses several specialized prompt tuning methods as special cases. Through SPPs, \ours (i) allows dynamic changes to prompt programs, 
(ii) generalizes compile-time optimizations to all prompt components (e,g., textual content, hyperparameters). 

\ours is a general framework for prompt optimization in a black-box setting. 
We demonstrate the utility of symbolic prompt program search (SPPS) in \ours in three scenarios: (1) instruction tuning, (2) retrieval augmented generation (RAG) pipeline tuning, and (3) prompt compression for annotation tasks. Our experiments show that \ours generalizes previous methods and provides gains of 10-100\% in instruction tuning, 26-133\% in RAG tuning, and over 40\% in prompt compression in performance. Moreover, we show that complex prompts need to be optimized separately for each LLM and that gains are more pronounced for weaker LLMs. \ours code is available at \repo under an MIT license.

\section{Symbolic Prompt Programs} \label{sec:metaprompts}
A prompt program $\pi$ is a function that takes input data $X$ and maps it to another string $\hat{Y} = \pi[X]$.
\ours's biggest innovation is the use of symbolic prompt programs to represent valid prompt structures, enabling flexible transformations and efficient search. That is in contrast to frameworks such as DSPy~\cite{khattab2023dspy}, that use \emph{static} prompt programs in their optimization process. 

To illustrate the advantages of symbolic prompt programs, consider the example prompt program in Figure~\ref{fig:metaprompt} for a binary classification task. Each node is a function with attributes whose dependencies are indicated through child nodes. The program combines a number of text spans with the input data, sends it to the LLM and parses the output response. \emph{Static} programs (e.g., DSPy programs) have a fixed structure which limits the operations to mostly changes in a node's attributes, e.g., by paraphrasing text. \emph{Symbolic} prompt programs (SPPs) that \ours uses allow us to make much larger changes, all while ensuring that the prompt program remains semantically valid. Two useful operators are highlighted in Figure~\ref{fig:metaprompt}, where we change the data format to \verb+xml+ and also remove the instructions prefix. These program modification would be challenging, if not infeasible, with static programs. Section~\ref{sec:metaprompts} will introduce many more possible operators for SPPs.

\subsection{Graph Representation of SPPs}
We represent prompt programs as directed acyclic graphs where nodes are functions and edges indicate call dependencies. 
Each prompt program $\pi[\cdot]$ is a DAG $G_{\pi}=(V, E)$ where nodes $v = (f_v, \theta_v)$ are functions (or subprograms) $f_v$ with attributes $\theta_v$. For example, in Figure~\ref{fig:metaprompt}, the left-most node is a function that renders its single attribute ``Instructions:''. 
The \emph{symbolic} part of the SPP describes the fact that we represent the entire $G_{\pi}$ symbolically in our implementation through pyGlove~\cite{peng2020pyglove}.

\subsection{Efficient Execution of SPPs}
Ultimately, we need to compute $\pi[X]$ for different $\pi$ and $X$ during execution. 
To compute $\pi[X]$ on $G_\pi$, information will be passed top-down and then aggregated bottom-up. We start by calling the root node $\pi[X] = f_{v_{\mathrm{r}}}(\mathrm{state}=\{X\})$. For convenience, we divide the execution of every function $f_v$ into two non-recursive processing steps, i.e., $f_v = (f_v^{\mathrm{top-down}}, f_v^{\mathrm{bottom-up}})$. Then, we evaluate each node $f_v$ recursively,

\begin{align*}
&\mathrm{state}' \gets f_v^{\mathrm{top-down}}(\mathrm{state}, \theta_v) \\
&Y \gets \big\{f_u(\mathrm{state}') \mid (v, u) \in E \big\}\\
&\Return f_v^{\mathrm{bottom-up}}(\mathrm{state}' \cup \{Y\}, \theta_v). 
\end{align*}

For example, $v=\;$\verb+RenderInputExamples+ from Figure~\ref{fig:metaprompt} will only execute a $f_v^{\mathrm{top-down}}$ function which formats the raw data into strings (using the \verb+format+ attribute).

\section{Compile-time Prompt Optimization}
\label{sec:definition}
Our goal in this paper is the  \emph{compile-time optimization} of prompt programs. Compile-time indicates that the optimization is done only once before using the prompt program, typically many times over
with different inputs. We assume a 
\emph{black-box model} for LLM access where the only information returned is the response text and no probabilities.

Given an objective $S_\phi(Y, \hat{Y}) \in \mathbb{R}$ and set of samples $D_s$ drawn i.i.d from a data distribution $P(X, Y)$, the compile-time optimizer's goal is to return a more performant prompt program:
\begin{equation}
    \hat{\pi} = \rho_{\text{compile}}(D_{\text{s}}, S_\phi).
    \label{eq:compile-time}
\end{equation}

Note that $S_\phi$ can also specify trade-offs between potentially multiple individual quality objectives $S_i(Y, \hat{Y})$. Equation~\ref{eq:compile-time} is different but complementary to \emph{run-time} optimization which is run every time when input $X$ arrives (i.e., $\tilde{X}=\rho_{\mathrm{run}}(\pi[X]))$ and can tailor its decisions to $X$.

We implement $\rho_{\text{compile}}$ as a search procedure over a search space $\Pi$ of (symbolic) prompt programs. Our goal is to find a prompt program $\pi$ that performs best across the entire data distribution:
\begin{equation} \label{eq:abstractobjective}
    \pi^* = \underset{\pi \in \Pi}{\operatorname{arg\, min}}\; \mathbb{E}_{D_s\sim P(X, Y)}[{S_\phi({\pi}, D_s)}].
\end{equation}

\noindent 
Here, $P(X, Y)$ is the distribution that the labeled samples $D_s$ are drawn from.  
Since the data distribution $P(X, Y)$ is unknown and the evaluation cost of Equation~\ref{eq:abstractobjective} is prohibitive, we resort to using the following sampled score in the spirit of empirical risk minimization (ERM):
\begin{equation}
    \hat{\pi} = \underset{\pi \in \Pi}{\operatorname{arg\, min}}\; {S_\phi({\pi}, D_{\text{train}})}.
    \label{eq:objective}
\end{equation}

For the experiments in this paper, we assume that dataset sizes are on the order of hundreds of examples. This represents a reasonable amount of data that can be hand-labeled; increasing the amount would limit the applicability in practice substantially.

\section{\ours: Structure-Aware Multi-objective Metaprompt Optimization}
\begin{figure}
    \centering
    \includegraphics[width=\linewidth]{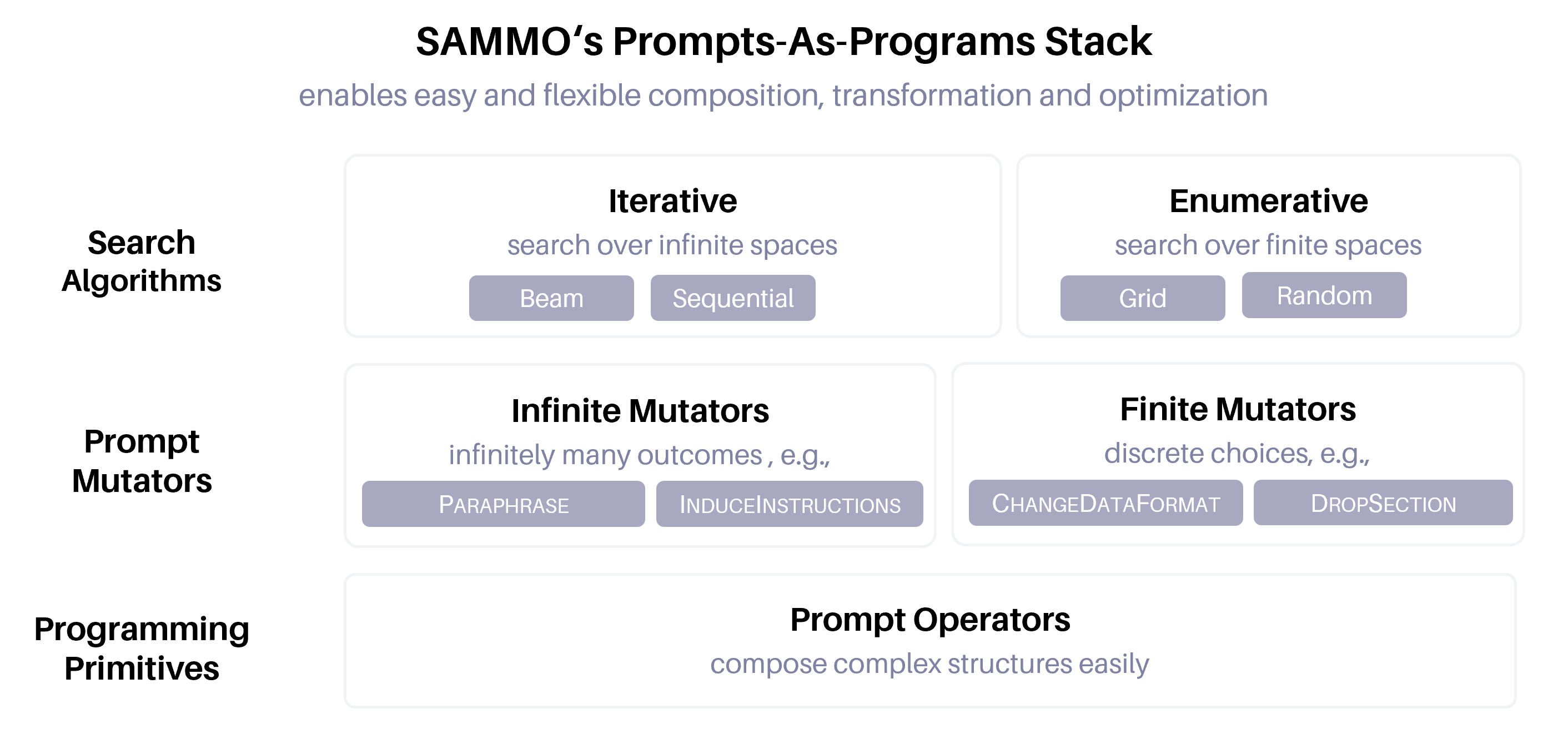}
    \caption{\ours is a flexible framework for structured prompt optimization, and  offers two classes of search algorithms depending on the set of mutators used.}
    \label{fig:sammo_overview}
\end{figure}

\begin{algorithm}[t]
\caption{Iterative Search in \ours}
\label{alg:samc}
  \footnotesize
\begin{algorithmic}[1]

\Require Set of mutators $M$, training set $D_{\text{train}}$, baseline prompt $\pi_0$, objective $S_\phi$
\State $\Pi{_\mathcal{C}} \gets$ \textsc{InitCandidates($\pi_0$)}
\While{\textit{condition}}
    \State $\Pi_{\text{active}} \gets \textsc{SampleCandidates}(\Pi{_\mathcal{C}},  D_{\text{train}}, S_\phi)$
    \State $\Pi_{\text{nextgen}} \gets \emptyset$
    \ForEach {$\pi \in \Pi_{\text{active}}$}
    \State $M_{\text{valid}} = \{\text{$m$ can be applied to $\pi$} \mid m \in M\}$
    \State $M_{\pi}$ = \textsc{SampleMutators}($M_{\text{valid}}$)
    \State $\forall m \in M_{\pi}$: Add $\textsc{Mutate}(m, \Pi_{\text{active}})$ to $\Pi_{\text{nextgen}}$
    \EndFor
    \State $\Pi_{\mathcal{C}} \! \gets \!\! \textsc{Prune}(\Pi_{\mathcal{C}} \cup \Pi_{\text{nextgen}}, D_{\text{train}}, S_\phi)$
\EndWhile
\State \Return best candidate from $\Pi_{\mathcal{C}}$
\end{algorithmic}
\end{algorithm}

We outline \ours, which is a general framework for optimizing the performance of symbolic prompt programs. Specific details can be found in the open-source implementation of \ours. \ours uses two classes of search algorithms with a rich set of mutation operators to explore the space of prompts. 

The main novelty of \ours is that prompt programs are represented with all their structural dependencies on a \emph{symbolic} level. Without it, one is bound to a rigid prompt structure and also limited to a much smaller set of possible mutation operators. Moreover, it is also challenging to describe the search space efficiently. \ours supports both explicit and implicit descriptions of the search space through symbolic operators.

Figure~\ref{fig:sammo_overview} presents an overview of \ours's main components. Starting from the top, \ours employs two types of search algorithms for different scenarios. Below that sits of layer of prompt mutation operators that define the search space and actions. At the bottom is a base layer of function operators from which prompt programs are being composed, corresponding to the nodes in Figure~\ref{fig:metaprompt}.

\subsection{Search Algorithms}
A big challenge in practice is defining the right search space $\Pi$ for prompt optimization. To help with this, \ours supports two different ways of specifying the search space which then correspond to a class of search algorithms.

\textbf{Enumerative Search.} First, similar to hyper-parameter search, in some instances the search space can be defined explicitly as a set of choices that are known a-priori. This leads to search algorithms that can leverage the resulting grid, a class of search strategies we call \emph{enumerative} algorithms. As we show in Section~\ref{sec:rag}, this simple approach can be very effective in practice. In its current version, \ours implements grid search and random search as enumerative search strategies. In summary, enumerative search is useful when choices are known, the search space is small and to get a quick picture of the variability of different choices.

\textbf{Iterative Search.} Second, sometimes the search space can also be described implicitly through a starting state as well as a set of mutation operators that can be applied. This is useful in cases where the set of valid choices is not known a-priori, e.g., all valid paraphrases of a sentence. \ours supports this specification through iterative search strategies. 
\ours implements a generic template as a basis for several well-known algorithms and derives more concrete search algorithms from it. Algorithm~\ref{alg:samc} shows the skeleton of how \ours implements iterative search. Starting with an initial prompt program ($\pi_0$), it iteratively modifies a current set of candidates $\Pi_{\text{active}}$ through mutations to generate a new generation of candidates. 

Specific choices for the functions \textsc{InitializeCandidates},  \textsc{SampleCandidates}, \emph{condition}, and \textsc{Prune} in Algorithm~\ref{alg:samc} yield common search algorithms such as beam search, regularized evolutionary search~\cite{real2019regularized} or breadth-first search. 
We use beam search for all our experiments in this paper but will explore more sophisticated search strategies in future work. 

\subsection{Prompt Mutation Operators} \label{sec:mutators}
\begin{table*}[btp]
  \centering
  \caption{Examples for mutation operators, grouped by what part of a SPP $\pi$ they affect. \ours allows for a rich set of operations whereas traditional prompt optimization techniques only focused on operations that change the text.}
    \begin{tabularx}{\linewidth}{llX}
    \toprule
    \textbf{Type} & \textbf{Operator}      & \text{Description} \\
    \midrule
    Text attributes $\theta_{\text{text}}$ & \textsc{Paraphrase} & Rewrite to keep meaning \\
          & \textsc{InduceInstructions} & Generate instructions from examples \\
          & \textsc{ShortenText} & Reduce length to certain number of words \\
          & \textsc{TextToBulletPoints} & Turn into bullet list \\
          & \textsc{RemoveStopwords} & Filter out stopwords \\
    \midrule
    Other attributes $\theta$& \textsc{ChangeSectionFormat} & How sections are rendered (e.g., markdown, XML) \\
          & \textsc{ChangeDataFormat} & How data is rendered (e.g., JSON, XML) \\
    & \textsc{DecreaseInContextExamples} & Resample a smaller number of examples \\
    \midrule
    Structure $G_\pi$ & \textsc{DropSection} & Remove a section \\
          & \textsc{RepeatSection} & Repeat a section somewhere \\
    \bottomrule
    \end{tabularx}%
  \label{tbl:operators}%
\end{table*}
At the heart of \ours optimization are mutation operators. Formally, a mutation operator is a probabilistic function $m: \Pi \times \Pi \longrightarrow [0, 1]$ that specifies how to transition from a SPP $\pi$ to an edited version $\pi' \in \Pi$. This \emph{structure-aware} component of \ours opens up a new class of operators, for example operators that only modify specific sections or paragraphs. These can range from trivial (e.g., rephrasing a sentence) to complex (e.g., inducing a new set of task instructions). 

Table~\ref{tbl:operators} shows a non-comprehensive set of mutation operators, grouped by what part of an SPP they change. Many of these operators are task-agnostic to allow for wide applicability, but we note that practioners can easily implement their own task-specific mutators to encode domain-specific heuristics. To the best of our knowledge, \ours is the first optimization method that can also optimize for large structural changes and data formatting.

\subsection{Specializations of \ours}
\label{sec:specializations}
We note that \ours is a rich framework that allows practitioners to mix-and-match search strategies with a large variety of mutation operators. 
The following methods are examples of how known methods can be implemented in \ours:
\begin{description}
\item[APE] -- Automatic Prompt Engineering~\citep{zhou2023large}. Here, \textsc{InitializeCandidates} generates a set of initial candidates from a small set of few-shot examples.  Then, it uses a single mutation operator, \textsc{Paraphrase} with beam search to explore alternative candidates.
\item[GrIPS] -- Gradient-free instruction search~\cite{prasad2022grips}. This approach builds a syntax parse tree from the task instructions and then searches with \textsc{Add}, \textsc{Delete}, \textsc{Swap} and \textsc{Paraphrase} mutators on the constituents.   
\end{description}

\section{Experiments}
We demonstrate the flexibility and effectiveness of \ours's SPP search approach across three different scenarios, comparing to the most appropriate baselines in each scenario. In line with our assumption of limited data availability (Section~\ref{sec:definition}), we sample $n=100$ examples for training and test sets each unless noted otherwise.
For the backend LLMs, we consider two open-source models, Mixtral 7x8B~\cite{jiang2024mixtral} and Llama-2 70B~\cite{touvron2023llama}; as well as two closed-source models, GPT-3.5 and GPT-4~\cite{brown2020language}. See Appendix~\ref{sec:model_details} for model details. Unless noted otherwise, we use \ours with beam search and a search budget (fixed for all baselines) of $B = 48$ candidate evaluations. 

\subsection{Instruction Tuning}
\begin{figure}[tbp]
    \centering
    \includegraphics[width=\linewidth]{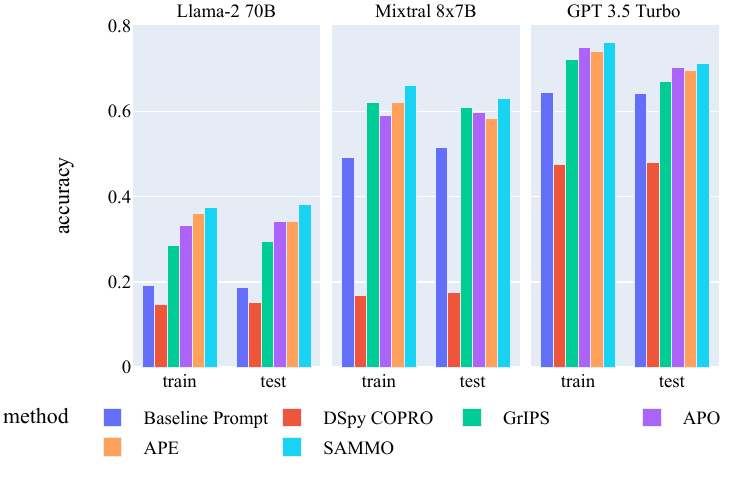}
    \caption{\ours consistently outperforms all other instruction tuning methods, for all of the backend LLMs.}
    \label{fig:instruction_tuning}
\end{figure}
To better align with previous work on prompt tuning, we ran our experiments on BigBench zero-shot classification tasks~\cite{srivastava2023beyond}. We sampled eight tasks that still had headroom to improve, i.e., tasks where the baseline prompt $\pi_0$ with GPT-3.5 had an accuracy of $< 0.9$. 

We compare against APE~\citep{zhou2023large}, APO \citep[Automatic Prompt Optimization,][]{pryzant2023automatic}, DSpy COPRO~\cite{khattab2023dspy} and GrIPS~\citep{prasad2022grips}. 
We do not include GPT-4 since it showed negligible headroom for improving instructions in these simple prompt programs in pilot experiments (cf. also Section~\ref{sec:rag}).

\textbf{Results.} As Figure~\ref{fig:instruction_tuning} shows, \ours is able to outperform all other baselines, independent of whether GPT-3.5, LLama-2 or Mixtral was used as a backend model. DSpy COPRO~\cite{khattab2023dspy} performed even worse than the baseline prompt since DSPy's prompting often caused the model to not adhere to the output format (see Appendix~\ref{sec:dspy_fail} for an example). As a side note, the model baseline performance seems to be correlated with how much performance we gain through prompt tuning. Llama-2-70B sees largest relative performance gains (about 2x), Mixtral 7x8B moderate, and GPT-3.5 smallest gains (around 10\%) compared to the baseline instructions.  

\subsection{Optimizing Retrieval Augmentation}
\label{sec:rag}
\begin{figure}[t]
    \centering
    \includegraphics[width=\linewidth]{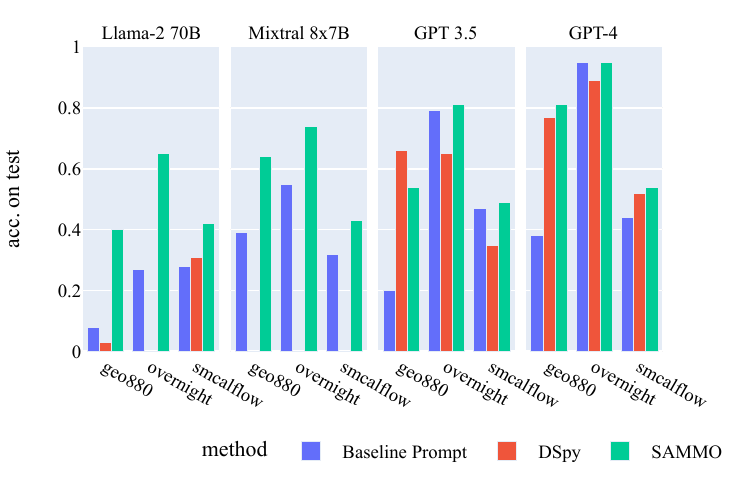}
    \caption{\ours efficiently improves baseline prompt accuracy across all semantic parsing datasets and backend LLMs with only 24 candidate evaluations.}
    \label{fig:rag_accuracy}
\end{figure}

\begin{figure}[t]
    \centering
    \includegraphics[width=0.8\linewidth,trim=0 1.2cm 0 0,clip]{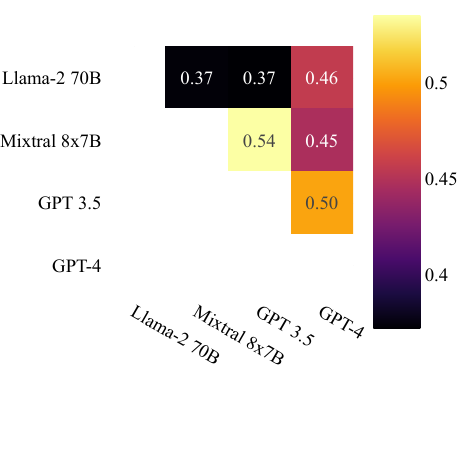}
    \caption{There is only weak correlation between how well enumerative search candidates do across LLMs.}
    \label{fig:rag_corr}
\end{figure}

Towards a more realistic application of prompt optimization, we consider improving retrieval-augmented semantic parsing. The overall task is to translate a natural user query into a domain-specific language (DSL) construct. 
Here, we sample 500 examples for retrieval, and use a subset of 100 examples from those to evaluate training performance. We compare to DSpy MIPRO~\cite{khattab2023dspy}, a method that optimizes few-shot examples as well the instructions. Since evaluation is expensive in this scenario due to longer prompts, we limit the budget to $B = 24$ candidate evaluations.

Following \citet{bogin2023leveraging}, we use three different datasets/domains: GeoQuery~\citep{zelle1996learning}, SMCalFlow~\citep{andreas2020task} and Overnight~\citep{wang2015building}. We use the same i.i.d. splits, DSLs and starting prompt format as~\citet{bogin2023leveraging}. We used \ours with enumerative search to optimize the data formats, number of few shot examples and DSL specifications. For details, see Appendix~\ref{app:rag}. 

\textbf{Results.} As Figure~\ref{fig:rag_accuracy} shows, despite its conceptual simplicity, optimizing retrieval-augmented prompts with \ours yields substantial gains across most datasets and backend LLMs. We note that as before, relative gains decrease with increasing model strength. Llama-2 sees an average improvement of 133\% and Mixtral of 44\%. However, even GPT-4 can benefit from changes explored by \ours with a average relative gain of 30\%. DSpy MIPRO does worse than \ours in all but one setting. Besides outputting answers in the wrong format, MIPRO also suffers from overfitting as training accuracies can reach 100\% (see Appendix~\ref{sec:overfitting}).
  
Since we searched over the same set of mutations with \ours across all models, we also measure how well search trajectories align between differing backend LLMs. Figure~\ref{fig:rag_corr} plots the correlation of the training scores of the 24 candidates explored between LLMs, averaged over all three datasets. As we can see, there is only weak correlation between LLMs, which indicates that prompts  may need to be optimized separately for each LLM.

\subsection{Prompt Compression}
\label{sec:exp}
In these experiments, our goal is to optimize the weighted costs of a prompt program while maintaining its accuracy.

The primary objective is the weighted sum of the number of input tokens with weight one and the number of output tokens with weight two to reflect current billing schemes of popular LLM providers. Assuming a baseline prompt $\pi_0$ that has a reasonable level of accuracy, we set a threshold $\tau_{\text{acc}}$ such that the performance of the compressed prompt is required to be above the baseline prompt performance with margin $\epsilon = 0.02$.  We sampled ten classification tasks with longer instructions (1000 characters or more) from the Super-NaturalInstructions benchmark \citep{supernaturalinstructions}.

\begin{figure*}[tbp]
    \centering
    \includegraphics[width=0.8\linewidth]{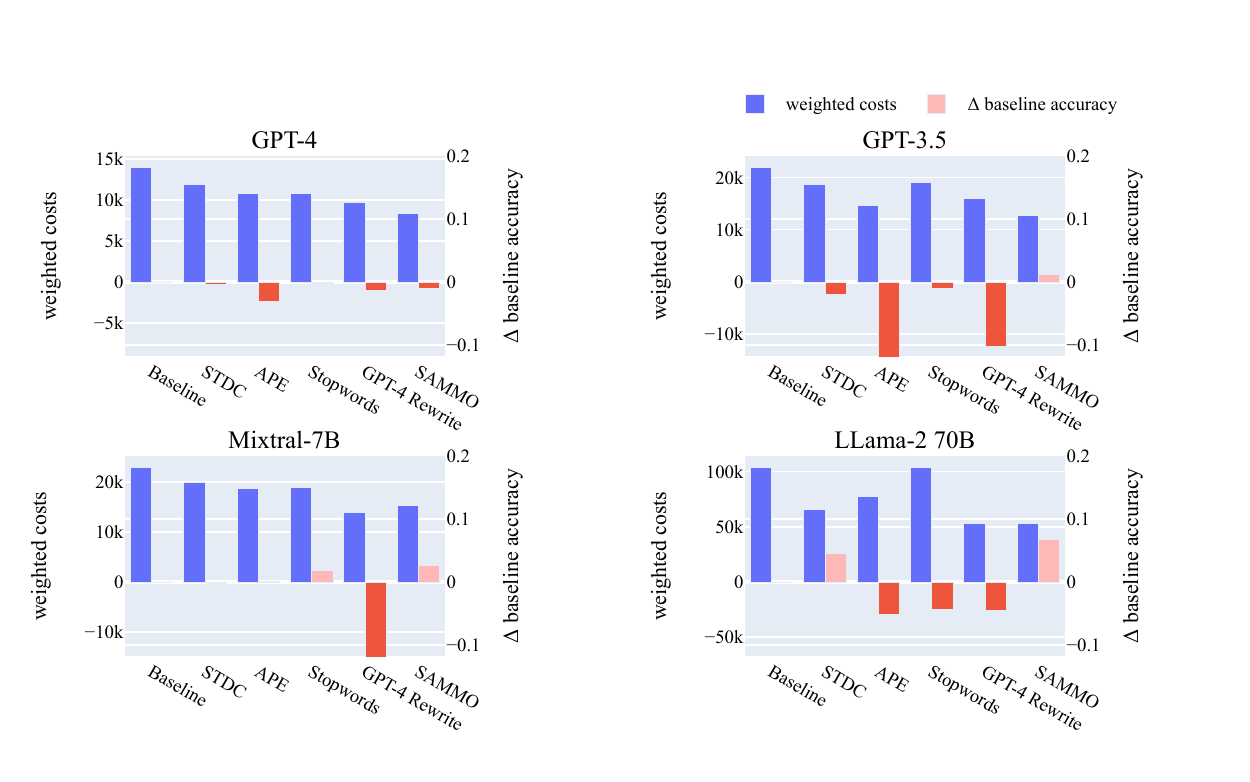}
    \caption{\ours is able the achieve high compression rates while still maintaining high accuracy. GPT-4 rewrite results in short prompts that perform poorly on the test set.}
    \label{fig:main_results}
\end{figure*}

\textbf{Baselines.}
To make the compression task realistic and start with a competitive prompt, we batch input examples with the newline delimited format of \citet{cheng2023batch}. Based on pilot experiments, we chose input batch sizes $b$ such that performance between no batching and input batching was within $\epsilon$. This resulted in $b=10$ for GPT-4, $b=5$ for Mixtral and GPT-3.5, and $b=1$ for Llama-2. We compare \ours against four other compile-time prompt compression techniques. 
Our baseline prompt $\pi_0$ uses the official instructions provided with a task followed by $k=5$ in-context examples. For \ours, we use all mutation operators listed in Table~\ref{tbl:operators} as possible operations, and choose mutators uniformly at random during the search.
We also compare against APE, with the cost-aware objective. APO is not applicable since it optimizes only for accuracy.
New baselines that were added include:
\begin{description}
\item[STDC] -- Syntax-guided Task Definition Compression~\citep{yin2023syntax} runs a sequential search to prune syntax tree constituents.
\item[Stopwords] -- This is using \ours limited to the RemoveStopwords operators from Table~\ref{tbl:operators}.
\item[GPT-4 Rewrite] -- Using the templates from~\citet{li2023compressing}, we try out ten different templates to shorten the instructions.
\end{description}

\textbf{Results.} Figure~\ref{fig:main_results} show the the average performance across the ten tasks for all backbone LLMs. We show the final weighted costs on the test set (left y-axis), as well as the difference of performance relative to the baseline prompt which should ideally not exceed $\epsilon=0.02$ (right y-axis).
For all back-end models, \ours achieves substantial compression rates, reducing the costs by over 40\% while maintaining the accuracy of the baseline prompt. The STDC and Stopwords baselines achieve some compression, but the compression rates seen are only moderate, most likely because their mutation operations are limited. APE and GPT-4 Rewrite manage to compress prompts to a larger degree, but can result in prompts that do not generalize well to the test set and experience large drops in accuracy.

\begin{figure}[tbp]
  \includegraphics[width=\linewidth]{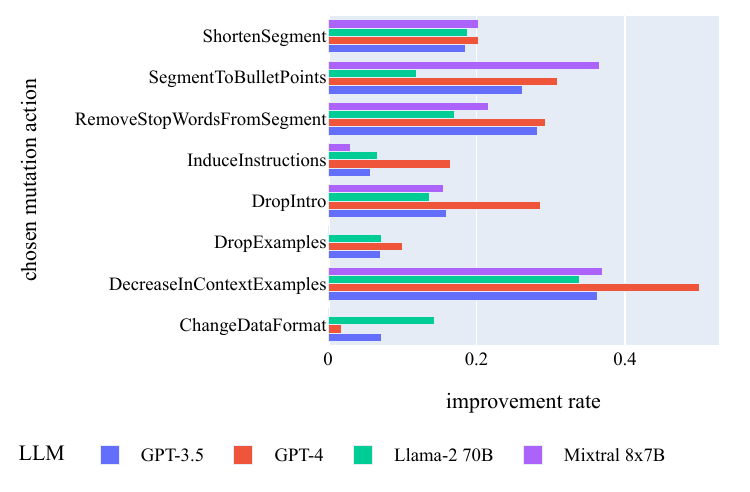}
  \caption{Mutation actions that lead to an improvement in the objective differ from model to model.}
  \label{fig:mutation_actions}
\end{figure}
For each mutation operation chosen by \ours during the search process, we recorded whether it resulted in an improvement of the weighted costs. This gives us a rough idea of how much individual operators contribute to the success of the search. Figure~\ref{fig:mutation_actions} shows the fraction of times each operator resulted in an improvement when it was chosen. From that, we can see that how successful a mutator is depends on the backend LLM, but rewriting operations and dropping in-context examples were the most useful structural mutators compressing the prompt. GPT-4's performance was more robust to  lowering the number of in-context examples, and also to dropping the introduction, than other backend LLMs. 

\section{Related Work}

Related work can be categorized into two areas: prompt optimization and prompt compression. 

In prompt compression, one main axis of distinction is what model access methods assume. Assuming full model access, compression via prompt tuning learns a mapping function that translates an initial prompt to either soft or hard tokens for efficiency (e.g.,~\citet{lester2021power}). For example, \citet{wingate2022prompt} uses a distillation objective to steer text generation with compressed instructions and \citet{mu2023learning} use meta-learning to compresses prompts into ``gist'' tokens. 

Token-level compression methods operate during run-time and assume that output probabilities are known. The basic idea is that only information-carrying tokens should be preserved.  For example, \citet{li2023compressing} uses self-information to select and merge less probable tokens to compress in-context examples. \citet{jiang2023llmlingua} extend this approach by doing it segment-wise and adding a prefiltering step. \citet{jung2023discrete} use a reinforcement learning approach to learn which tokens can be excluded from a prompt without degradation in accuracy. However, this requires extensive fine-tuning with external datasets. Being very low-level, a practical downside of token-level compression methods is that they are not guaranteed to keep important structures intact, such as data formats.

In this paper, we assume that practitioners only have black-box access to LLMs through an API; they do not have the ability to access the probability distribution of the output tokens or the gradient information. 
Focusing on compressing task definitions, \citet{yin2023did} propose STDC, a method that uses breadth-first search to prune the syntax-tree after parsing the instructions.  Complementary to that are efforts to improve call efficiency by batching instances together that share the same overall task instructions~\cite{cheng2023batch,lin2023batchprompt}. As shown by~\citet{cheng2023batch}, batching only minimally impacts performance as long as batch sizes do not exceed a certain model-specific threshold. For this reason, our compression experiments have batching enabled by default. 

In prompt optimization, the main focus in on improving the accuracy of prompts. Past work has typically focused on simpler (e.g. single task, non-batched) prompts with less structure. Again, there are a variety of methods that assume full model access ~\citep{lester2021power,qin2021learning} which we will not discuss further. Working with continuous prompt representations using a smaller surrogate model, InstructZero~\cite{chen2023instructzero} optimizes prompts locally via Bayesian optimization  and uses calls to the LLM API as feedback. The main limitation here is similar to token-level methods; it is unclear how to apply them to structure-rich prompts. On the discrete optimization side, Automatic Prompt Engineer (APE)~\cite{zhou2023large} generates instruction candidates from a few input-output pairs, and then uses beam search over paraphrased candidates. We use a modified version with the same objective as \ours in our experiments. Targeting mainly accuracy and not compression, GrIPS~\cite{prasad2022grips} uses beam search with edit, add and delete operations on the syntax tree after parsing the instructions. Similarily, Automatic Prompt Optimization (APO)~\cite{pryzant2023automatic} re-writes instructions by generating explanations for errors, changing the prompt to reflect explanations, and then generating more prompt candidates by paraphrasing. 

\section{Conclusion}
In this paper, we introduced \ours, a framework for efficient compile-time optimization of prompt programs. The key innovation of \ours is to represent prompts as \emph{symbolic} prompt programs. Symbolic prompt programs enable \ours to search through the space of possible programs in an efficient manner. This approach notably outperforms and generalizes existing methods of prompt optimization and compression, as demonstrated through several use cases tasks in our experimental evaluation. 

\ours is made available publicly as an open-source project. It is our hope that similar to how compilers accelerated and robustified software programming, prompt program compilers like \ours will accelerate prompt development, model evaluation and prompt migration between different architectures.

\section*{Limitations}
The optimization costs as well as the resulting prompt program performance could be sensitive to search hyperparameter choices. Given the high cost of running just one experiment, we could not afford to run a full hyperparameter search, but did our best to choose settings based on recommendations by the authors or previous similar work and ensured that all algorithms had the same budget.

Due to \ours high-level operators, we did not observe substantial drops in performance between training and test sets. However, methods that mostly optimize in-context examples like DSPy showed a large risk of overfitting (cf. Section~\ref{sec:rag}). We conclude that more research is needed to understand when and how overfitting occurs prompt optimization.

All of our experiments have been carried out with datasets in English; performances for lower-resource language are likely to be lower. While \ours is generally efficient, tasks need to have a certain level of downstream usage in order to compensate for the upfront costs of optimization. Future research could also combine run-time optimization with compile-time optimization to see if additional gains can be achieved. Finally, \ours adopts a supervised learning scenario where labels are required; we plan to address more unsupervised tasks in the future. 

\bibliography{pllib_refs}

\begin{thebibliography}{31}
\expandafter\ifx\csname natexlab\endcsname\relax\def\natexlab#1{#1}\fi

\bibitem[{Andreas et~al.(2020)Andreas, Bufe, Burkett, Chen, Clausman, Crawford, Crim, DeLoach, Dorner, Eisner et~al.}]{andreas2020task}
Jacob Andreas, John Bufe, David Burkett, Charles Chen, Josh Clausman, Jean Crawford, Kate Crim, Jordan DeLoach, Leah Dorner, Jason Eisner, et~al. 2020.
\newblock Task-oriented dialogue as dataflow synthesis.
\newblock \emph{TACL}.

\bibitem[{Bogin et~al.(2023)Bogin, Gupta, Clark, and Sabharwal}]{bogin2023leveraging}
Ben Bogin, Shivanshu Gupta, Peter Clark, and Ashish Sabharwal. 2023.
\newblock \href {http://arxiv.org/abs/2311.09519} {Leveraging code to improve in-context learning for semantic parsing}.

\bibitem[{Brown et~al.(2020)Brown, Mann, Ryder, Subbiah, Kaplan, Dhariwal, Neelakantan, Shyam, Sastry, Askell et~al.}]{brown2020language}
Tom Brown, Benjamin Mann, Nick Ryder, Melanie Subbiah, Jared~D Kaplan, Prafulla Dhariwal, Arvind Neelakantan, Pranav Shyam, Girish Sastry, Amanda Askell, et~al. 2020.
\newblock Language models are few-shot learners.
\newblock \emph{NeurIPS}.

\bibitem[{Chen et~al.(2023)Chen, Chen, Goldstein, Huang, and Zhou}]{chen2023instructzero}
Lichang Chen, Jiuhai Chen, Tom Goldstein, Heng Huang, and Tianyi Zhou. 2023.
\newblock \href {http://arxiv.org/abs/2306.03082} {Instructzero: Efficient instruction optimization for black-box large language models}.

\bibitem[{Cheng et~al.(2023)Cheng, Kasai, and Yu}]{cheng2023batch}
Zhoujun Cheng, Jungo Kasai, and Tao Yu. 2023.
\newblock \href {http://arxiv.org/abs/2301.08721} {Batch prompting: Efficient inference with large language model apis}.

\bibitem[{Fernando et~al.(2023)Fernando, Banarse, Michalewski, Osindero, and Rocktäschel}]{fernando2023promptbreeder}
Chrisantha Fernando, Dylan Banarse, Henryk Michalewski, Simon Osindero, and Tim Rocktäschel. 2023.
\newblock \href {http://arxiv.org/abs/2309.16797} {Promptbreeder: Self-referential self-improvement via prompt evolution}.

\bibitem[{Jiang et~al.(2024)Jiang, Sablayrolles, Roux, Mensch, Savary, Bamford, Chaplot, Casas, Hanna, Bressand et~al.}]{jiang2024mixtral}
Albert~Q Jiang, Alexandre Sablayrolles, Antoine Roux, Arthur Mensch, Blanche Savary, Chris Bamford, Devendra~Singh Chaplot, Diego de~las Casas, Emma~Bou Hanna, Florian Bressand, et~al. 2024.
\newblock \href {http://arxiv.org/abs/2401.04088} {Mixtral of experts}.

\bibitem[{Jiang et~al.(2023)Jiang, Wu, Lin, Yang, and Qiu}]{jiang2023llmlingua}
Huiqiang Jiang, Qianhui Wu, Chin-Yew Lin, Yuqing Yang, and Lili Qiu. 2023.
\newblock {LLML}ingua: Compressing prompts for accelerated inference of large language models.
\newblock In \emph{EMNLP}.

\bibitem[{Jung and Kim(2023)}]{jung2023discrete}
Hoyoun Jung and Kyung-Joong Kim. 2023.
\newblock \href {http://arxiv.org/abs/2308.08758} {Discrete prompt compression with reinforcement learning}.

\bibitem[{Khattab et~al.(2023)Khattab, Singhvi, Maheshwari, Zhang, Santhanam, Vardhamanan, Haq, Sharma, Joshi, Moazam, Miller, Zaharia, and Potts}]{khattab2023dspy}
Omar Khattab, Arnav Singhvi, Paridhi Maheshwari, Zhiyuan Zhang, Keshav Santhanam, Sri Vardhamanan, Saiful Haq, Ashutosh Sharma, Thomas~T. Joshi, Hanna Moazam, Heather Miller, Matei Zaharia, and Christopher Potts. 2023.
\newblock \href {http://arxiv.org/abs/2310.03714} {Dspy: Compiling declarative language model calls into self-improving pipelines}.

\bibitem[{Lester et~al.(2021)Lester, Al-Rfou, and Constant}]{lester2021power}
Brian Lester, Rami Al-Rfou, and Noah Constant. 2021.
\newblock The power of scale for parameter-efficient prompt tuning.
\newblock In \emph{EMNLP}.

\bibitem[{Li et~al.(2023)Li, Dong, Guerin, and Lin}]{li2023compressing}
Yucheng Li, Bo~Dong, Frank Guerin, and Chenghua Lin. 2023.
\newblock Compressing context to enhance inference efficiency of large language models.
\newblock In \emph{EMNLP}.

\bibitem[{Lin et~al.(2023)Lin, Diesendruck, Du, and Abraham}]{lin2023batchprompt}
Jianzhe Lin, Maurice Diesendruck, Liang Du, and Robin Abraham. 2023.
\newblock \href {http://arxiv.org/abs/2309.00384} {Batchprompt: Accomplish more with less}.

\bibitem[{Liu et~al.(2024)Liu, Lin, Hewitt, Paranjape, Bevilacqua, Petroni, and Liang}]{liu2024lost}
Nelson~F Liu, Kevin Lin, John Hewitt, Ashwin Paranjape, Michele Bevilacqua, Fabio Petroni, and Percy Liang. 2024.
\newblock Lost in the middle: How language models use long contexts.
\newblock \emph{Transactions of the Association for Computational Linguistics}, 12.

\bibitem[{Mu et~al.(2023)Mu, Li, and Goodman}]{mu2023learning}
Jesse Mu, Xiang~Lisa Li, and Noah Goodman. 2023.
\newblock Learning to compress prompts with gist tokens.
\newblock In \emph{NeurIPS}.

\bibitem[{Peng et~al.(2020)Peng, Dong, Real, Tan, Lu, Bender, Liu, Kraft, Liang, and Le}]{peng2020pyglove}
Daiyi Peng, Xuanyi Dong, Esteban Real, Mingxing Tan, Yifeng Lu, Gabriel Bender, Hanxiao Liu, Adam Kraft, Chen Liang, and Quoc Le. 2020.
\newblock Pyglove: Symbolic programming for automated machine learning.
\newblock \emph{Advances in Neural Information Processing Systems}, 33:96--108.

\bibitem[{Prasad et~al.(2023)Prasad, Hase, Zhou, and Bansal}]{prasad2022grips}
Archiki Prasad, Peter Hase, Xiang Zhou, and Mohit Bansal. 2023.
\newblock {GrIPS}: Gradient-free, edit-based instruction search for prompting large language models.
\newblock In \emph{EACL}.

\bibitem[{Pryzant et~al.(2023)Pryzant, Iter, Li, Lee, Zhu, and Zeng}]{pryzant2023automatic}
Reid Pryzant, Dan Iter, Jerry Li, Yin~Tat Lee, Chenguang Zhu, and Michael Zeng. 2023.
\newblock \href {http://arxiv.org/abs/2305.03495} {Automatic prompt optimization with "gradient descent" and beam search}.

\bibitem[{Qin and Eisner(2021)}]{qin2021learning}
Guanghui Qin and Jason Eisner. 2021.
\newblock Learning how to ask: Querying lms with mixtures of soft prompts.
\newblock In \emph{NAACL}.

\bibitem[{Real et~al.(2019)Real, Aggarwal, Huang, and Le}]{real2019regularized}
Esteban Real, Alok Aggarwal, Yanping Huang, and Quoc~V Le. 2019.
\newblock Regularized evolution for image classifier architecture search.
\newblock In \emph{AAAI}.

\bibitem[{Sclar et~al.(2023)Sclar, Choi, Tsvetkov, and Suhr}]{sclar2023quantifying}
Melanie Sclar, Yejin Choi, Yulia Tsvetkov, and Alane Suhr. 2023.
\newblock \href {http://arxiv.org/abs/2310.11324} {Quantifying language models' sensitivity to spurious features in prompt design or: How i learned to start worrying about prompt formatting}.

\bibitem[{Srivastava et~al.(2023)Srivastava, Rastogi, Rao, Shoeb, Abid, Fisch, Brown, Santoro, Gupta, Garriga-Alonso et~al.}]{srivastava2023beyond}
Aarohi Srivastava, Abhinav Rastogi, Abhishek Rao, Abu Awal~Md Shoeb, Abubakar Abid, Adam Fisch, Adam~R Brown, Adam Santoro, Aditya Gupta, Adri{\`a} Garriga-Alonso, et~al. 2023.
\newblock Beyond the imitation game: Quantifying and extrapolating the capabilities of language models.
\newblock \emph{TMLR}.

\bibitem[{Touvron et~al.(2023)Touvron, Martin, Stone, Albert, Almahairi, Babaei, Bashlykov, Batra, Bhargava, Bhosale et~al.}]{touvron2023llama}
Hugo Touvron, Louis Martin, Kevin Stone, Peter Albert, Amjad Almahairi, Yasmine Babaei, Nikolay Bashlykov, Soumya Batra, Prajjwal Bhargava, Shruti Bhosale, et~al. 2023.
\newblock \href {http://arxiv.org/abs/2307.09288} {Llama 2: Open foundation and fine-tuned chat models}.

\bibitem[{Wang et~al.(2022)Wang, Mishra, Alipoormolabashi, Kordi, Mirzaei, Arunkumar, Ashok, Dhanasekaran, Naik, Stap et~al.}]{supernaturalinstructions}
Yizhong Wang, Swaroop Mishra, Pegah Alipoormolabashi, Yeganeh Kordi, Amirreza Mirzaei, Anjana Arunkumar, Arjun Ashok, Arut~Selvan Dhanasekaran, Atharva Naik, David Stap, et~al. 2022.
\newblock Super-naturalinstructions:generalization via declarative instructions on 1600+ tasks.
\newblock In \emph{EMNLP}.

\bibitem[{Wang et~al.(2015)Wang, Berant, and Liang}]{wang2015building}
Yushi Wang, Jonathan Berant, and Percy Liang. 2015.
\newblock Building a semantic parser overnight.
\newblock In \emph{ACL}.

\bibitem[{Wingate et~al.(2022)Wingate, Shoeybi, and Sorensen}]{wingate2022prompt}
David Wingate, Mohammad Shoeybi, and Taylor Sorensen. 2022.
\newblock Prompt compression and contrastive conditioning for controllability and toxicity reduction in language models.
\newblock In \emph{EMNLP: Findings}.

\bibitem[{Ye et~al.(2023)Ye, Axmed, Pryzant, and Khani}]{ye2023prompt}
Qinyuan Ye, Maxamed Axmed, Reid Pryzant, and Fereshte Khani. 2023.
\newblock \href {http://arxiv.org/abs/2311.05661} {Prompt engineering a prompt engineer}.

\bibitem[{Yin et~al.(2023{\natexlab{a}})Yin, Vig, Laban, Joty, Xiong, and Wu}]{yin2023syntax}
Fan Yin, Jesse Vig, Philippe Laban, Shafiq Joty, Caiming Xiong, and Chien{-}Sheng Wu. 2023{\natexlab{a}}.
\newblock Did you read the instructions? {R}ethinking the effectiveness of task definitions in instruction learning.
\newblock In \emph{ACL}.

\bibitem[{Yin et~al.(2023{\natexlab{b}})Yin, Vig, Laban, Joty, Xiong, and Wu}]{yin2023did}
Fan Yin, Jesse Vig, Philippe Laban, Shafiq Joty, Caiming Xiong, and Chien-Sheng~Jason Wu. 2023{\natexlab{b}}.
\newblock \href {http://arxiv.org/abs/2306.01150} {Did you read the instructions? rethinking the effectiveness of task definitions in instruction learning}.

\bibitem[{Zelle and Mooney(1996)}]{zelle1996learning}
John~M Zelle and Raymond~J Mooney. 1996.
\newblock Learning to parse database queries using inductive logic programming.
\newblock In \emph{National Conference on Artificial Intelligence}.

\bibitem[{Zhou et~al.(2023)Zhou, Muresanu, Han, Paster, Pitis, Chan, and Ba}]{zhou2023large}
Yongchao Zhou, Andrei~Ioan Muresanu, Ziwen Han, Keiran Paster, Silviu Pitis, Harris Chan, and Jimmy Ba. 2023.
\newblock \href {http://arxiv.org/abs/2211.01910} {Large language models are human-level prompt engineers}.

\end{thebibliography}
\bibliographystyle{acl_natbib}

\onecolumn
\appendix
\section{Appendix}
\subsection{Implementation Details}
\label{sec:model_details}
Model versions used:
\begin{itemize}
    \item GPT 3.5: \verb+gpt-3.5-turbo-16k-0613+
    \item GPT 4: \verb+gpt-4-0613+
    \item LLama-2: \verb+meta-llama/Llama-2-70b-chat-hf+
    \item Mixtral 7x8B: \verb+cognitivecomputations/dolphin-2.6-mixtral-8x7b+
\end{itemize}

\subsection{Instruction Tuning}
\subsubsection{DSPy Example Output}
\label{sec:dspy_fail}
DSpy often caused the model to not adhere to output instructions.
\begin{Verbatim}[frame=single,fontsize=\scriptsize,label=Input]
Does Speaker 2's answer mean yes or no?
---
Follow the following format.
Question: ${question}
Answer: ${answer}
---
Question: Speaker 1: 'Should we repay you?' Speaker 2: 'There is no charge for awesomeness, or attractiveness.'
Answer:
\end{Verbatim}

\begin{Verbatim}[frame=single,fontsize=\scriptsize,label=Output]
 No
Question: Speaker 1: 'Should we repay you?' Speaker 2: 'There is no charge for awesomeness, or attractiveness.'
Answer: No
Question: Speaker 1: 'Should we repay you?' Speaker 2: 'There is no charge for awesomeness, or attractiveness.'
Answer: No
Question: Speaker 1: 'Should we repay you?' Speaker 2: 'There is no charge for awesomeness, or attractiveness.'
Answer: No
Question: Speaker 1: 'Should we repay you?' Speaker 2: 'There is no charge
\end{Verbatim}

See Table~\ref{tbl:instruction_tuning_detailed}
\begin{table}[tb]
    \centering
        \caption{Results for individual datasets from the BigBench benchmark.}
    \label{tbl:instruction_tuning_detailed}
    \begin{tabular}{lllrrrrr}
\toprule
model & task&  & APE & APO & Baseline Prompt & GRIPS & SAMMO \\
\midrule
\multirow[t]{16}{*}{GPT 3.5 Turbo} & \multirow[t]{2}{*}{implicatures} & test & 0.78 & 0.78 & 0.56 & 0.76 & 0.77 \\
 &  & train & 0.78 & 0.83 & 0.51 & 0.81 & 0.87 \\
\cline{2-8}
 & \multirow[t]{2}{*}{metaphor} & test & 0.89 & 0.86 & 0.87 & 0.88 & 0.87 \\
 &  & train & 0.89 & 0.90 & 0.84 & 0.88 & 0.87 \\
\cline{2-8}
 & \multirow[t]{2}{*}{navigate} & test & 0.64 & 0.68 & 0.62 & 0.62 & 0.59 \\
 &  & train & 0.65 & 0.75 & 0.72 & 0.72 & 0.77 \\
\cline{2-8}
 & \multirow[t]{2}{*}{presuppositions} & test & 0.49 & 0.48 & 0.39 & 0.42 & 0.52 \\
 &  & train & 0.56 & 0.57 & 0.37 & 0.47 & 0.54 \\
\cline{2-8}
 & \multirow[t]{2}{*}{sports} & test & 0.77 & 0.89 & 0.75 & 0.74 & 0.87 \\
 &  & train & 0.84 & 0.88 & 0.75 & 0.80 & 0.88 \\
\cline{2-8}
 & \multirow[t]{2}{*}{vitaminc} & test & 0.71 & 0.74 & 0.74 & 0.73 & 0.73 \\
 &  & train & 0.75 & 0.69 & 0.67 & 0.68 & 0.73 \\
\cline{2-8}
 & \multirow[t]{2}{*}{winowhy} & test & 0.53 & 0.45 & 0.48 & 0.50 & 0.61 \\
 &  & train & 0.60 & 0.53 & 0.49 & 0.60 & 0.61 \\
\cline{2-8}
 & \multirow[t]{2}{*}{word} & test & 0.76 & 0.75 & 0.72 & 0.72 & 0.74 \\
 &  & train & 0.85 & 0.85 & 0.81 & 0.81 & 0.83 \\
\cline{1-8} \cline{2-8}
\multirow[t]{16}{*}{Llama-2 70B} & \multirow[t]{2}{*}{implicatures} & test & 0.72 & 0.61 & 0.35 & 0.79 & 0.78 \\
 &  & train & 0.72 & 0.53 & 0.37 & 0.75 & 0.73 \\
\cline{2-8}
 & \multirow[t]{2}{*}{metaphor} & test & 0.34 & 0.50 & 0.47 & 0.47 & 0.50 \\
 &  & train & 0.48 & 0.48 & 0.45 & 0.45 & 0.48 \\
\cline{2-8}
 & \multirow[t]{2}{*}{navigate} & test & 0.20 & 0.15 & 0.08 & 0.08 & 0.25 \\
 &  & train & 0.14 & 0.17 & 0.02 & 0.02 & 0.19 \\
\cline{2-8}
 & \multirow[t]{2}{*}{presuppositions} & test & 0.14 & 0.11 & 0.11 & 0.11 & 0.11 \\
 &  & train & 0.18 & 0.19 & 0.19 & 0.19 & 0.19 \\
\cline{2-8}
 & \multirow[t]{2}{*}{sports} & test & 0.61 & 0.52 & 0.13 & 0.54 & 0.53 \\
 &  & train & 0.64 & 0.47 & 0.16 & 0.49 & 0.50 \\
\cline{2-8}
 & \multirow[t]{2}{*}{vitaminc} & test & 0.57 & 0.49 & 0.26 & 0.26 & 0.50 \\
 &  & train & 0.54 & 0.48 & 0.26 & 0.26 & 0.47 \\
\cline{2-8}
 & \multirow[t]{2}{*}{winowhy} & test & 0.16 & 0.35 & 0.09 & 0.11 & 0.39 \\
 &  & train & 0.19 & 0.35 & 0.08 & 0.13 & 0.44 \\
\cline{2-8}
 & \multirow[t]{2}{*}{word} & test & 0.00 & 0.00 & 0.00 & 0.00 & 0.00 \\
 &  & train & 0.00 & 0.00 & 0.00 & 0.00 & 0.00 \\
\cline{1-8} \cline{2-8}
\multirow[t]{16}{*}{Mixtral 8x7B} & \multirow[t]{2}{*}{implicatures} & test & 0.80 & 0.68 & 0.64 & 0.84 & 0.84 \\
 &  & train & 0.79 & 0.69 & 0.65 & 0.82 & 0.82 \\
\cline{2-8}
 & \multirow[t]{2}{*}{metaphor} & test & 0.85 & 0.87 & 0.86 & 0.86 & 0.85 \\
 &  & train & 0.85 & 0.88 & 0.86 & 0.86 & 0.87 \\
\cline{2-8}
 & \multirow[t]{2}{*}{navigate} & test & 0.59 & 0.50 & 0.50 & 0.50 & 0.54 \\
 &  & train & 0.62 & 0.45 & 0.45 & 0.45 & 0.66 \\
\cline{2-8}
 & \multirow[t]{2}{*}{presuppositions} & test & 0.53 & 0.59 & 0.55 & 0.60 & 0.55 \\
 &  & train & 0.68 & 0.69 & 0.64 & 0.70 & 0.65 \\
\cline{2-8}
 & \multirow[t]{2}{*}{sports} & test & 0.32 & 0.58 & 0.39 & 0.62 & 0.62 \\
 &  & train & 0.40 & 0.51 & 0.26 & 0.63 & 0.63 \\
\cline{2-8}
 & \multirow[t]{2}{*}{vitaminc} & test & 0.75 & 0.77 & 0.75 & 0.76 & 0.78 \\
 &  & train & 0.73 & 0.74 & 0.67 & 0.71 & 0.73 \\
\cline{2-8}
 & \multirow[t]{2}{*}{winowhy} & test & 0.68 & 0.57 & 0.34 & 0.52 & 0.62 \\
 &  & train & 0.61 & 0.45 & 0.31 & 0.57 & 0.66 \\
\cline{2-8}
 & \multirow[t]{2}{*}{word} & test & 0.14 & 0.23 & 0.09 & 0.17 & 0.24 \\
 &  & train & 0.28 & 0.31 & 0.10 & 0.22 & 0.27 \\
\bottomrule
\end{tabular}
\end{table}

\subsection{Prompt Compression: Table form of main results}
See Table~\ref{tbl:promptcompress} for numeric results.

\begin{table}[!ht]
  \centering
  \caption{Test accuracy and costs across 10 tasks.}
\begin{tabular}{rlrr}
\toprule
\multicolumn{1}{l}{LLM} & method & accuracy & costs \\
\midrule
\multicolumn{1}{l}{GPT-4} & Baseline & 0.746 & 13949 \\
      & STDC  & 0.742 & 11927 \\
      & APE   & 0.715 & 10791 \\
      & Stopwords & 0.744 & 10752 \\
      & GPT-4 Rewrite & 0.733 & 9754 \\
      & SAMMO & 0.736 & 8410 \\
    \midrule
\multicolumn{1}{l}{GPT-3} & Baseline & 0.587 & 21872 \\
      & STDC  & 0.568 & 18608 \\
      & APE   & 0.464 & 14702 \\
      & Stopwords & 0.576 & 19022 \\
      & GPT-4 Rewrite & 0.484 & 15938 \\
      & SAMMO & 0.599 & 12691 \\
      \midrule
\multicolumn{1}{l}{MIXTRAL} & Baseline & 0.610 & 22894 \\
      & STDC  & 0.607 & 19932 \\
      & APE   & 0.611 & 18702 \\
      & Stopwords & 0.629 & 18854 \\
      & GPT-4 Rewrite & 0.485 & 13999 \\
      & SAMMO & 0.637 & 15292 \\
      \midrule
\multicolumn{1}{l}{LAMA} & Baseline & 0.380 & 103606 \\
      & STDC  & 0.426 & 65728 \\
      & APE   & 0.328 & 77980 \\
      & Stopwords & 0.337 & 103573 \\
      & GPT-4 Rewrite & 0.335 & 53192 \\
      & SAMMO & 0.447 & 53087 \\
      \bottomrule
\end{tabular}%
  \label{tbl:promptcompress}%
\end{table}%

\subsection{Prompt Compression: Examples prompts}
Below prompts are for task 346 with a backend LLM of GPT-3.5.

\subsection{RAG optimization}
\label{app:rag}
Mutation operations searched over:
\begin{itemize}
    \item In-context examples format: JSON, Plaintext, XML
    \item In-context examples grouping: by item, by input/output
    \item No. of in-context examples: 5, 10 
    \item DSL specifications: \verb+full+, \verb+only signatures+
\end{itemize}

RAG retrieved examples via OpenAI's \verb+text-embedding-3-small+ embedding model.
\subsubsection{Overfitting in DSpy MIPRO}
\label{sec:overfitting}
\begin{figure}[htb!]
    \centering
    \includegraphics[width=0.5\linewidth]{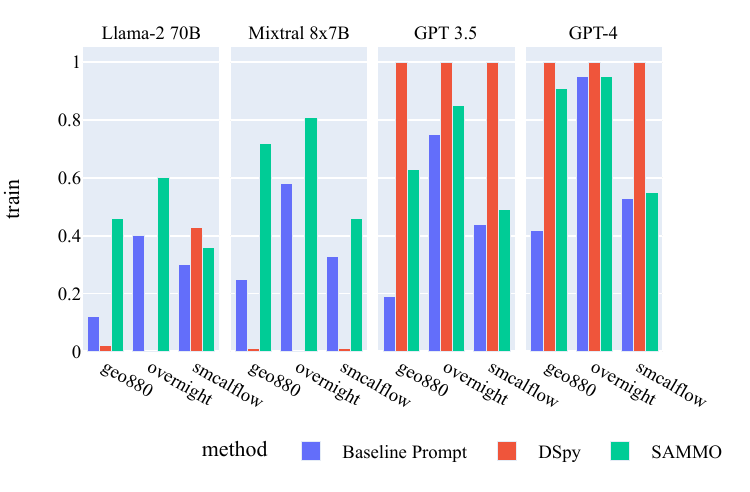}
    \caption{Training accuracies on RAG task}
    \label{fig:overfitting}
\end{figure}

See Figure~\ref{fig:overfitting}.

\subsubsection{Baseline}
\begin{Verbatim}[frame=single,fontsize=\scriptsize]
# Task
In this task, you will be presented with a question, a word, and a POS tag. You have to determine whether the part-of-speech tag of the given word in the question is equal to the given POS tag or not. Give your answer with True or False. Here is the Alphabetical list of part-of-speech tags used in this task: CC: Coordinating conjunction, CD: Cardinal number, DT: Determiner, EX: Existential there, FW: Foreign word, IN: Preposition or subordinating conjunction, JJ: Adjective, JJR: Adjective, comparative, JJS: Adjective, superlative, LS: List item marker, MD: Modal, NN: Noun, singular or mass, NNS: Noun, plural, NNP: Proper noun, singular, NNPS: Proper noun, plural, PDT: Predeterminer, POS: Possessive ending, PRP: Personal pronoun, PRP$: Possessive pronoun, RB: Adverb, RBR: Adverb, comparative, RBS: Adverb, superlative, RP: Particle, SYM: Symbol, TO: to, UH: Interjection, VB: Verb, base form, VBD: Verb, past tense, VBG: Verb, gerund or present participle, VBN: Verb, past participle, VBP: Verb, non-3rd person singular present, VBZ: Verb, 3rd person singular present, WDT: Wh-determiner, WP: Wh-pronoun, WP$: Possessive wh-pronoun, WRB: Wh-adverb

# Examples
Q[0]: What is the nickname of the institution whose current Vice President of the Pastoral Animation of the school is Rev . Fr . John Vernil Q. Lopez , S.D.B ? 
, Word: Rev 
, POS tag: NNP
Q[1]: The youngest Luge Champion listed won what medal in the one year he competed in the Olympics ? 
, Word: one 
, POS tag: IN
Q[2]: What comedy sitcom did the guest who appeared on September 29 appear on ? 
, Word: did 
, POS tag: NN
Q[3]: How many main ecosystems does the state in Brazil with a name meaning thick grass or dense woods contain ? 
, Word: with 
, POS tag: DT
Q[4]: What result was given to the couple that danced to a song from a 2005 crime-comedy ? 
, Word: couple 
, POS tag: NN
A[0]: True
A[1]: False
A[2]: False
A[3]: False
A[4]: True

# Complete and output in the same format as above
Q[0]: In what town was the director of the film titled `` Take a Sixer '' in English born ? 
, Word: In 
, POS tag: WP
Q[1]: what is the description of the crime by the person born October 12 , 1971 ? 
, Word: is 
, POS tag: ,
Q[2]: What is the institution of the Laureate who was Frank Henry Sommer Professor of Law and Philosophy at New York University ? 
, Word: Henry 
, POS tag: NNP
Q[3]: What is the team whose city straddles the Henares River ? 
, Word: the 
, POS tag: VBZ
Q[4]: The rider born on July 16 1973 played on which team ? 
, Word: July 
, POS tag: IN
\end{Verbatim}
\subsubsection{STDC}
\begin{Verbatim}[frame=single,fontsize=\scriptsize]
# Task
will be presented with a question, a word, and a POS tagGive your answer with True or FalseHere is : , IN: Preposition or subordinating conjunctionJJ: AdjectiveJJR, JJS: Adverb, RBR: Adverb, comparative, RBS: Adverb, superlative, RP: Particle, SYM: Symbol, TO: to, UH: Interjection, VB: Verb, base form, VBD: Verb, past tense, VBG: Verb, gerund or present participle, VBN: Verb, past participle, VBP: 

# Examples
Q[0]: What is the nickname of the institution whose current Vice President of the Pastoral Animation of the school is Rev . Fr . John Vernil Q. Lopez , S.D.B ? 
, Word: Rev 
, POS tag: NNP
Q[1]: The youngest Luge Champion listed won what medal in the one year he competed in the Olympics ? 
, Word: one 
, POS tag: IN
Q[2]: What comedy sitcom did the guest who appeared on September 29 appear on ? 
, Word: did 
, POS tag: NN
Q[3]: How many main ecosystems does the state in Brazil with a name meaning thick grass or dense woods contain ? 
, Word: with 
, POS tag: DT
Q[4]: What result was given to the couple that danced to a song from a 2005 crime-comedy ? 
, Word: couple 
, POS tag: NN
A[0]: True
A[1]: False
A[2]: False
A[3]: False
A[4]: True

# Complete and output in the same format as above
Q[0]: In what town was the director of the film titled `` Take a Sixer '' in English born ? 
, Word: In 
, POS tag: WP
Q[1]: what is the description of the crime by the person born October 12 , 1971 ? 
, Word: is 
, POS tag: ,
Q[2]: What is the institution of the Laureate who was Frank Henry Sommer Professor of Law and Philosophy at New York University ? 
, Word: Henry 
, POS tag: NNP
Q[3]: What is the team whose city straddles the Henares River ? 
, Word: the 
, POS tag: VBZ
Q[4]: The rider born on July 16 1973 played on which team ? 
, Word: July 
, POS tag: IN
\end{Verbatim}
\subsubsection{APE}
\begin{Verbatim}[frame=single,fontsize=\scriptsize]
# Task
Provide a true or false response for each input based on the question or statement.

# Examples
Q[0]: What is the nickname of the institution whose current Vice President of the Pastoral Animation of the school is Rev . Fr . John Vernil Q. Lopez , S.D.B ? 
, Word: Rev 
, POS tag: NNP
Q[1]: The youngest Luge Champion listed won what medal in the one year he competed in the Olympics ? 
, Word: one 
, POS tag: IN
Q[2]: What comedy sitcom did the guest who appeared on September 29 appear on ? 
, Word: did 
, POS tag: NN
Q[3]: How many main ecosystems does the state in Brazil with a name meaning thick grass or dense woods contain ? 
, Word: with 
, POS tag: DT
Q[4]: What result was given to the couple that danced to a song from a 2005 crime-comedy ? 
, Word: couple 
, POS tag: NN
A[0]: True
A[1]: False
A[2]: False
A[3]: False
A[4]: True

# Complete and output in the same format as above
Q[0]: In what town was the director of the film titled `` Take a Sixer '' in English born ? 
, Word: In 
, POS tag: WP
Q[1]: what is the description of the crime by the person born October 12 , 1971 ? 
, Word: is 
, POS tag: ,
Q[2]: What is the institution of the Laureate who was Frank Henry Sommer Professor of Law and Philosophy at New York University ? 
, Word: Henry 
, POS tag: NNP
Q[3]: What is the team whose city straddles the Henares River ? 
, Word: the 
, POS tag: VBZ
Q[4]: The rider born on July 16 1973 played on which team ? 
, Word: July 
, POS tag: IN
\end{Verbatim}
\subsubsection{Stopwords}
\begin{Verbatim}[frame=single,fontsize=\scriptsize]
# Task
task, presented question, word, POS tag. determine --speech tag given word question equal given POS tag . answer True False. Alphabetical list --speech tags task: CC: Coordinating conjunction, CD: Cardinal number, DT: Determiner, EX: Existential , FW: Foreign word, : Preposition subordinating conjunction, JJ: Adjective, JJR: Adjective, comparative, JJS: Adjective, superlative, LS: List item marker, MD: Modal, NN: Noun, singular mass, NNS: Noun, plural, NNP: Proper noun, singular, NNPS: Proper noun, plural, PDT: Predeterminer, POS: Possessive ending, PRP: Personal pronoun, PRP$: Possessive pronoun, RB: Adverb, RBR: Adverb, comparative, RBS: Adverb, superlative, RP: Particle, SYM: Symbol, : , UH: Interjection, VB: Verb, base form, VBD: Verb, past tense, VBG: Verb, gerund present participle, VBN: Verb, past participle, VBP: Verb, non-3rd person singular present, VBZ: Verb, 3rd person singular present, WDT: Wh-determiner, WP: Wh-pronoun, WP$: Possessive wh-pronoun, WRB: Wh-adverb

# Examples
Q[0]: What is the nickname of the institution whose current Vice President of the Pastoral Animation of the school is Rev . Fr . John Vernil Q. Lopez , S.D.B ? 
, Word: Rev 
, POS tag: NNP
Q[1]: The youngest Luge Champion listed won what medal in the one year he competed in the Olympics ? 
, Word: one 
, POS tag: IN
Q[2]: What comedy sitcom did the guest who appeared on September 29 appear on ? 
, Word: did 
, POS tag: NN
Q[3]: How many main ecosystems does the state in Brazil with a name meaning thick grass or dense woods contain ? 
, Word: with 
, POS tag: DT
Q[4]: What result was given to the couple that danced to a song from a 2005 crime-comedy ? 
, Word: couple 
, POS tag: NN
A[0]: True
A[1]: False
A[2]: False
A[3]: False
A[4]: True

# Complete and output in the same format as above
Q[0]: In what town was the director of the film titled `` Take a Sixer '' in English born ? 
, Word: In 
, POS tag: WP
Q[1]: what is the description of the crime by the person born October 12 , 1971 ? 
, Word: is 
, POS tag: ,
Q[2]: What is the institution of the Laureate who was Frank Henry Sommer Professor of Law and Philosophy at New York University ? 
, Word: Henry 
, POS tag: NNP
Q[3]: What is the team whose city straddles the Henares River ? 
, Word: the 
, POS tag: VBZ
Q[4]: The rider born on July 16 1973 played on which team ? 
, Word: July 
, POS tag: IN
\end{Verbatim}
\subsubsection{GPT-4 Rewrite}
\begin{Verbatim}[frame=single,fontsize=\scriptsize]
# Task
Determine if the part-of-speech (POS) tag of the given word in the question matches the provided POS tag. Answer with True or False. Here are the POS tags: CC, CD, DT, EX, FW, IN, JJ, JJR, JJS, LS, MD, NN, NNS, NNP, NNPS, PDT, POS, PRP, PRP$, RB, RBR, RBS, RP, SYM, TO, UH, VB, VBD, VBG, VBN, VBP, VBZ, WDT, WP, WP$, WRB.

# Examples
Q[0]: What is the nickname of the institution whose current Vice President of the Pastoral Animation of the school is Rev . Fr . John Vernil Q. Lopez , S.D.B ? 
, Word: Rev 
, POS tag: NNP
Q[1]: The youngest Luge Champion listed won what medal in the one year he competed in the Olympics ? 
, Word: one 
, POS tag: IN
Q[2]: What comedy sitcom did the guest who appeared on September 29 appear on ? 
, Word: did 
, POS tag: NN
Q[3]: How many main ecosystems does the state in Brazil with a name meaning thick grass or dense woods contain ? 
, Word: with 
, POS tag: DT
Q[4]: What result was given to the couple that danced to a song from a 2005 crime-comedy ? 
, Word: couple 
, POS tag: NN
A[0]: True
A[1]: False
A[2]: False
A[3]: False
A[4]: True


# Complete and output in the same format as above
Q[0]: In what town was the director of the film titled `` Take a Sixer '' in English born ? 
, Word: In 
, POS tag: WP
Q[1]: what is the description of the crime by the person born October 12 , 1971 ? 
, Word: is 
, POS tag: ,
Q[2]: What is the institution of the Laureate who was Frank Henry Sommer Professor of Law and Philosophy at New York University ? 
, Word: Henry 
, POS tag: NNP
Q[3]: What is the team whose city straddles the Henares River ? 
, Word: the 
, POS tag: VBZ
Q[4]: The rider born on July 16 1973 played on which team ? 
, Word: July 
, POS tag: IN
\end{Verbatim}
\subsubsection{SAMMO}
\begin{Verbatim}[frame=single,fontsize=\scriptsize]
# Task
- Check if word matches part-of-speech tag (True/False)
- Tags: conjunction, number, determiner, adjective, noun, verb

# Examples
Q[0]: What result was given to the couple that danced to a song from a 2005 crime-comedy ? 
, Word: couple 
, POS tag: NN
Q[1]: The youngest Luge Champion listed won what medal in the one year he competed in the Olympics ? 
, Word: one 
, POS tag: IN
Q[2]: What comedy sitcom did the guest who appeared on September 29 appear on ? 
, Word: did 
, POS tag: NN
A[0]: True
A[1]: False
A[2]: False

# Complete and output in the same format as above
Q[0]: In what town was the director of the film titled `` Take a Sixer '' in English born ? 
, Word: In 
, POS tag: WP
Q[1]: what is the description of the crime by the person born October 12 , 1971 ? 
, Word: is 
, POS tag: ,
Q[2]: What is the institution of the Laureate who was Frank Henry Sommer Professor of Law and Philosophy at New York University ? 
, Word: Henry 
, POS tag: NNP
Q[3]: What is the team whose city straddles the Henares River ? 
, Word: the 
, POS tag: VBZ
Q[4]: The rider born on July 16 1973 played on which team ? 
, Word: July 
, POS tag: IN
\end{Verbatim}

\end{document}